%% file: main.tex
\documentclass[
]{ceurart}

\usepackage{listings}
\usepackage{amsmath}

\begin{document}

\copyrightyear{2026}
\copyrightclause{Copyright for this paper by its authors.
  Use permitted under Creative Commons License Attribution 4.0
  International (CC BY 4.0).}

\conference{TRUST-AI: The Second European Workshop on Trustworthy AI. Organized as part of the International Joint Conference on Artificial Intelligence - IJCAI/ECAI 2026. August 2026, Bremen, Germany.}

\title{Evaluating and Mitigating Gender Bias in Pre-trained Embeddings for ML-based Recruitment}

\author[1]{Farnaz Faramarzi Lighvan}[%
orcid=0009-0009-7505-7128,
email=Farnaz.faramarzi.lighvan@vub.be,
]
\cormark[1]

\author[1]{Lynn Houthuys}[%
orcid=0000-0003-0867-4168,
email=Lynn.houthuys@vub.be,
]
\address[1]{AI Lab, Vrije Universiteit Brussel, Belgium}

\cortext[1]{Corresponding author.}

\begin{abstract}
AI-based recruitment systems that rely on machine learning models trained on historical CV data, risk perpetuating and amplifying social biases. A key challenge arises in unstructured CV text, where pre-trained language model embeddings may infer sensitive attributes such as gender even after explicit indicators are removed. In this paper, we evaluate nine pre-trained embedding models on the synthetic FairCVdb dataset, analyzing the informativeness of their embeddings for applicant scoring and their susceptibility to gender leakage, on both original and gender-scrubbed biographies. We further use a multi-task adversarial learning framework with gradient reversal to predict applicant suitability while suppressing gender information from learned representations. Finally, we use a multi-objective Pareto-front-based model selection to balance predictive utility and fairness. Our experimental results show that explicit gender scrubbing substantially reduces but does not eliminate gender leakage, while adversarial learning improves fairness mainly on original biographies and acts as a complementary strategy rather than a substitute for text-level debiasing.
\end{abstract}


\maketitle

\section{Introduction}

Artificial intelligence is increasingly being adopted in recruitment systems to support candidate screening and hiring decisions, and reducing manual effort when evaluating large numbers of applicants \cite{AIreqruitment}. In many such systems, applicant CVs are used to predict suitability scores or generate hiring recommendations. However, these systems raise serious concerns about fairness and discrimination toward demographic groups such as defined by gender and ethnicity~\cite{ethical}, further reflected in regulation such as the EU AI Act~\cite{EUAIact}, classifying recruitment and candidate evaluation as high-risk and subjecting them to bias-mitigation, transparency, and human-oversight obligations.

AI is commonly used in recruitment in two ways. The first relies on large language models (LLMs) to match job descriptions with candidate CVs. Although efficient, LLM-based approaches have shown biased output generation \cite{LLMbias2} and can be less transparent and harder to control in high-stakes settings. Prior work has examined LLM-based resume matching with respect to gender, race, maternity and pregnancy status, political affiliation, and education \cite{LLMcategories}.

The second approach trains machine learning (ML) models on historical recruitment data \cite{MLreqruitment}. These models can be tailored to specific job contexts, but their targets may reflect existing social and organisational biases \cite{facebookdiscrimination}. In particular, suitability scores assigned by human decision-makers may encode bias, which ML models can reproduce or amplify \cite{biaschallenge}. A common mitigation strategy is to remove sensitive attributes from the input. While this is relatively straightforward for structured CV features, it is more difficult for unstructured text. For gender bias, names, pronouns, and titles can be removed or replaced with gender-neutral alternatives \cite{de2019bias,dictionary}, but remaining proxy words may still reveal gender \cite{de2019bias,semanticsname}. Amazon's discontinued recruitment model illustrates this risk, as it penalised resumes containing terms associated with female applicants after being trained on historically male-dominated hiring data \cite{dastin2022amazon}. Therefore, textual CV inputs often require careful scrubbing of explicit gendered terms, such as replacing ``mother'' and ``fireman'' with ``parent'' and ``firefighter'' \cite{petkova2025gender}.

For unstructured text, natural language processing methods encode CV text into vector representations. Pre-trained embedding models can capture rich semantic information for tasks such as applicant ranking or score prediction, but they may also infer gender from implicit cues such as word choice, writing style, and linguistic patterns. Petkova et al. \cite{petkova2025gender} proposed round-trip translation to reduce such implicit gender cues, while Pena et al. \cite{pena2025LLM} proposed explainability and adversarial learning as training-time alternatives for mitigating gender bias in LLM embeddings.

The main objective of this paper is to determine whether multi-task adversarial training can substitute for, or complement, preprocessing-stage gender scrubbing when pre-trained embeddings are used for recruitment scoring. To study this, we train multilayer perceptron (MLP) models on the synthetic FairCVdb dataset \cite{pena2023human}, which provides applicant CVs together with gender-biased and unbiased suitability scores. We deliberately train on gender-biased scores, since unbiased labels are rarely available in real-world recruitment; the unbiased scores are used only as a reference for ranking-based fairness evaluation. Using nine pre-trained embedding models on both original and gender-scrubbed biographies, we first train a model to predict the suitability score for a candidate, as well as a separate gender classification model to analyze gender leakage. We then use a multi-task adversarial framework with a gradient reversal layer \cite{grl} to jointly predict suitability and gender while suppressing gender-related information in the shared representation. The main contributions are:

\begin{itemize}
    \item A comparative analysis of gender-leakage potential across nine embedding models on original and gender-scrubbed biographies in a recruitment scoring setting.
    \item A mid-fusion neural network architecture that processes biography embeddings before combining them with structured CV features for increased flexibility for information fusion of both modalities.
    \item A multi-task adversarial learning framework with gradient reversal for suppressing gender information from shared representations while preserving scoring utility.
    \item A Pareto-front-based multi-objective optimization strategy for principled model selection under an explicit accuracy--fairness trade-off.
    \item An empirical investigation into whether adversarial training can substitute for, or complement preprocessing-stage gender scrubbing.
\end{itemize}
The most closely related work is the recent study by Pena et al. \cite{pena2025LLM}, which uses a similar multi-task framework for bias mitigation in ML-based recruitment. While this shows the potential of the approach, this study effectively extends on this in several key areas. More specifically: (1) by considering both original and scrubbed biographies and a more diverse set of embeddings, allowing for a more thorough analysis on the importance of scrubbing and selection of  embedding model.; (2) by using a Pareto-front-based model selection instead of using a fixed fairness penalty weight, allowing for an explicit and principled accuracy–fairness trade-off.

\section{Methodology}

This section describes the experimental pipeline used to study bias-aware score prediction on FairCVdb. The task is formulated as a regression problem in which each candidate is represented by structured CV attributes and a textual biography, and the model predicts a hiring suitability score in the interval $[0,1]$. Models are trained on gender-biased targets, reflecting realistic conditions where unbiased labels are unavailable; however, the goal is to achieve accurate score prediction while reducing unfair differences between male and female applicants.

We conduct three sets of experiments using the same mid-fusion strategy to combine the input modalities of the dataset. First, we train a single-task gender classifier on both original and gender-scrubbed biographies to measure the gender-leakage potential of each embedding model. Second, we train a single-task regressor for score prediction using both biography variants, evaluating the effect of text-level debiasing and the utility of each embedding model. Third, we use a multi-task adversarial neural network with shared fusion layers, task-specific output heads, and a gradient reversal layer to learn representations that are informative for score prediction while reducing gender information. For model selection, we use a Pareto-front-based strategy to balance predictive accuracy and fairness instead of fixing a fairness penalty weight manually.

\subsection{FairCVdb dataset}

We conduct all experiments on FairCVdb \cite{pena2020bias}, a synthetic dataset of 24,000 resume profiles designed to study fairness in automated recruitment. Each profile includes structured candidate information, demographic attributes, occupation information, a name, and a short biography. The structured features contain seven competency attributes derived from education, availability, previous experience, recommendation letter status, and language proficiency, together with a suitability attribute measuring the affinity between the candidate's occupation sector and the target job. The dataset is split into 19,200 training and 4,800 test profiles, balanced across gender, ethnicity, and occupational sector.

FairCVdb provides unbiased scores $T^U$, gender-biased scores $T^G$, and ethnicity-biased scores. In this work, we focus on gender bias and train all models using $T^G$, since real-world recruitment data would typically contain biased historical labels rather than unbiased ground truth. The unbiased scores $T^U$ are not used for training or model selection; they are reserved only for computing ranking-based fairness metrics during evaluation, as described in Section~\ref{metrics}.

The dataset provides two biography versions: original biographies, which include explicit gender indicators such as names, pronouns, and titles, and gender-scrubbed biographies, where these explicit indicators are removed. We use only the structured competency-related features and biography text, excluding face image embeddings.

\subsection{Embedding models}

We evaluate nine pre-trained text representation models spanning four model groups to encode the biography sections of the CVs.

\noindent\textbf{Static word embeddings.}
Static word embeddings assign one fixed vector to each word, independent of context. We include fastText~\cite{fasttext}(\texttt{crawl-300d-2M}) and GloVe~\cite{glove} (\texttt{glove.6B.300d}), both using 300-dimensional pre-trained word vectors. For each biography, token-level vectors are averaged to produce a fixed-size biography representation.

\noindent\textbf{Contextual transformer models.}
Contextual transformer models produce token representations that depend on the surrounding text. We include \texttt{bert-base-uncased} model of BERT \cite{bert}, \texttt{FacebookAI/roberta-base} model of RoBERTa~\cite{roberta}, and \texttt{microsoft/mpnet-base} of MPNet~\cite{mpnet}. Biography embeddings are obtained by mean-pooling the last hidden states over non-padding tokens, using a maximum sequence length of 256 tokens.

\noindent\textbf{Sentence transformer models.} Sentence transformer models are optimized to produce semantically meaningful sentence-level embeddings. We include \texttt{all-mpnet-base-v2}, a fine-tuned version of MPNet \cite{mpnet}, and \texttt{all-MiniLM-L6-v2}, a fine-tuned version of MiniLM \cite{minilm}, both adapted within the SentenceTransformers framework \cite{sentencebert} on large sentence-pair datasets using a contrastive objective, and optimised specifically for producing semantically meaningful fixed-size sentence representations.

\noindent\textbf{OpenAI API embeddings.} These are large-scale proprietary embedding models trained 
on broad web-scale corpora, being among the current 
state-of-the-art in dense text representation. We include \texttt{text-embedding-3-large} and \texttt{text-embedding-3-small}, accessed via the OpenAI API \cite{openai}. 
\subsection{Single-Task Gender Classification}

We train an MLP for binary gender classification using BCE loss. The input is the concatenation of standardized structured CV features and the biography embedding. Gender classification accuracy is used as a proxy for gender leakage: higher accuracy indicates that the representation retains more gender-revealing information. We run this experiment on both original and gender-scrubbed biographies to assess how much gender information remains after scrubbing.

\subsection{Single-Task Score Prediction}

We train an MLP regression model to predict the gender-biased hiring score $T^G$ using MAE loss. The input is the concatenation of standardized structured features and the biography embedding. Fairness is considered during model selection by tuning hyperparameters using the Pareto-front-based procedure in Section~\ref{pareto}. Experiments are run on both original and gender-scrubbed biographies to evaluate the effect of text debiasing on accuracy and fairness.

\subsection{Multi-Task Adversarial Learning with Gradient Reversal}

To reduce gender information in the learned representation, we employ a multi-task adversarial learning architecture with a Gradient Reversal Layer (GRL) \cite{grl}. The model consists of a shared encoder, a score prediction head, and an auxiliary gender prediction head.

\begin{figure}
    \centering
    \includegraphics[width=0.6\linewidth]{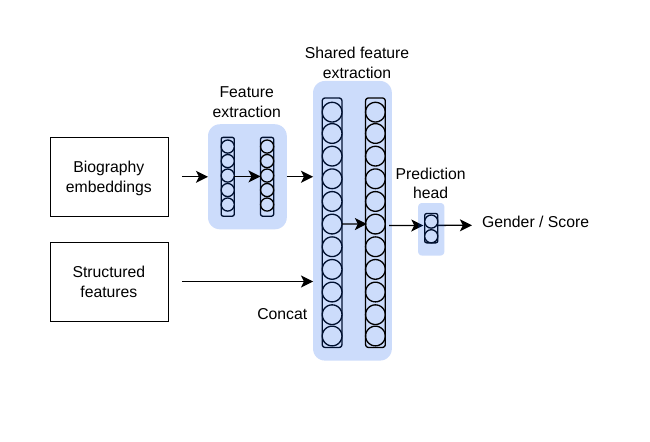}  \\[-1.2cm]
    \includegraphics[width=0.6\linewidth]{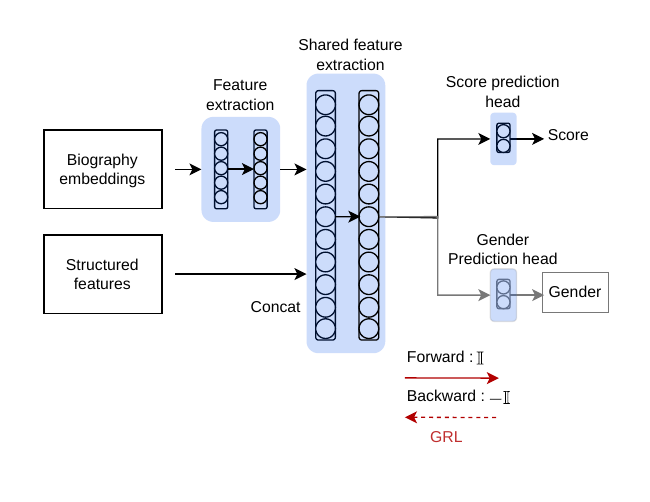} 
    \vspace{-0.5cm}
    \caption{Architectures used for single-task and multi-task experiments. Both use the same mid-fusion strategy: biography embeddings are processed by a feature extractor, concatenated with structured features, and passed through a shared feature extractor. (Top) Single-task architecture for gender classification and score prediction with a single prediction head. (Bottom) Multi-task adversarial architecture with a score prediction head and a gender prediction head connected through a gradient reversal layer (GRL), which acts as the identity in the forward pass and reverses the gender-head gradient during backpropagation.}
    \label{fig:architecture}
\end{figure}

Given the input representation $\mathbf{x}_i$, the shared encoder computes a latent representation:
\begin{equation}
\mathbf{z}_i = f_{\theta}(\mathbf{x}_i).
\end{equation}
The score prediction head $h_{\phi}$ predicts the hiring score, while the adversarial gender head $a_{\psi}$ predicts the gender attribute from the reversed representation:
\begin{equation}
\hat{y}_i = h_{\phi}(\mathbf{z}_i), \quad \quad \hat{g}_i = a_{\psi}(\mathrm{GRL}(\mathbf{z}_i)).
\end{equation}
The GRL acts as an identity function in the forward pass, but reverses the gradient during backpropagation:
\begin{equation}
\frac{\partial \mathrm{GRL}(\mathbf{z})}{\partial \mathbf{z}} = -\mathbf{I}.
\end{equation}
where $\mathbf{I}$ is identity matrix. The joint training objective is:
\begin{equation}
\mathcal{L} = \mathcal{L}_{\text{score}} + \lambda\, \mathcal{L}_{\text{gender}}.
\label{eq:mt_loss}
\end{equation}
Because of the gradient reversal, minimizing this objective updates the gender head to improve gender prediction, while simultaneously updating the shared encoder in the opposite direction, encouraging it to discard gender-predictive information from $\mathbf{z}$. The score prediction head is trained with MAE loss and the gender head with Binary Cross-Entropy (BCE) loss. The adversarial weight $\lambda$ is treated as a tunable hyperparameter. As with the single-task setting, both prediction accuracy and fairness are considered during hyperparameter tuning using the Pareto front-based procedure described in Section~\ref{pareto}, and experiments are conducted on both original and gender-scrubbed biographies across all nine embedding models.
Figure~\ref{fig:architecture} shows the single-task and multi-task architectures used in our experiments.
\subsection{Performance and Fairness Metrics}\label{metrics}
For score prediction, we report the following performance and fairness metrics.

\noindent\textbf{MAE} measures the average absolute deviation between predicted scores and gender-biased ground-truth scores $T^G$, reflecting predictive utility.  

\noindent\textbf{KL divergence.}
We use Kullback--Leibler (KL) divergence \cite{KL} to measure the distributional disparity between the predicted score distributions of male and female candidates. Lower values indicate that the male and female predicted-score distributions are more similar, suggesting greater distributional parity.

\noindent\textbf{P-percent rule (P\%)} measures demographic parity as the ratio of female-to-male (or male-to-female, whichever is smaller) representation in the top-100 predicted scores. A value of 1.0 indicates perfect parity; lower values indicate greater disparity.

\noindent\textbf{Recall@100 by gender.}
Using top-100 candidates from unbiased ground-truth scores $T^U$, we measure the fraction of these candidates that are recovered in the model's predicted top-100 list. At top 100  candidates, we report Male Recall, Female Recall, Average Recall, and Recall Gap, where Recall Gap is the absolute difference between male and female recall. Although $T^U$ would generally not be available in a real-world deployment, it is available in the synthetic FairCVdb dataset. This allows us to evaluate how well each model recovers truly qualified candidates across gender groups. A high Average Recall together with a small Recall Gap indicates a ranking model that is both accurate and more balanced across gender groups.

\subsection{Fairness-Aware Multi-Objective Optimization}
\label{pareto}

For both the single-task and multi-task settings, we use a fairness-aware validation objective during hyperparameter tuning:
\begin{equation}
\mathcal{J}_{\alpha}
=
\mathrm{MAE}_{\mathrm{val}}
+
\alpha D_{\mathrm{KL},\mathrm{val}},
\quad
\alpha \in \mathcal{A},
\end{equation}
where $\mathrm{MAE}_{\mathrm{val}}$ is the validation prediction error, $D_{\mathrm{KL},\mathrm{val}}$ is the KL divergence between male and female predicted score distributions on the validation set, and $\alpha$ controls the relative importance of distributional fairness during model selection. Rather than fixing a single $\alpha$ manually, we evaluate a grid of candidate values $\mathcal{A}$. For each $\alpha$, model hyperparameters are tuned independently via Optuna \cite{optuna}, producing one best validation configuration associated with a pair $(\mathrm{MAE}^{(\alpha)}_{\mathrm{val}}, D^{(\alpha)}_{\mathrm{KL},\mathrm{val}})$.

After all $\alpha$ values have been evaluated, we identify the Pareto front in the objective space of these pairs. A configuration is Pareto-efficient if no other configuration achieves both lower validation MAE and lower validation KL simultaneously, with at least one strictly improved. Because the MAE is computed against the gender-biased targets, it measures fidelity to biased labels rather than correctness with respect to any unbiased notion of suitability. The accuracy axis of the Pareto front should therefore not be read as predictive quality in an absolute sense: a configuration with a slightly higher MAE can be preferable, since some deviation from the biased labels is expected when gender-related information is suppressed.
To ensure that fairness gains are not achieved at an unreasonable deviation from the biased labels, we apply an MAE tolerance constraint. Let $\mathrm{MAE}^{(0)}_{\mathrm{val}}$ denote the validation MAE of the selected configuration when $\alpha=0$. We retain only Pareto-efficient configurations satisfying:
\begin{equation}
\mathrm{MAE}^{(\alpha)}_{\mathrm{val}}
\leq
(1+\delta)\,\mathrm{MAE}^{(0)}_{\mathrm{val}}.
\end{equation}
where $\delta$ is the acceptable MAE deviation threshold. Among the remaining candidates, we select the configuration with the lowest validation KL:
\begin{equation}
\alpha^{*}
=
\arg\min_{\alpha \in \mathcal{P}_{\delta}}
D_{\mathrm{KL},\mathrm{val}}^{(\alpha)},
\end{equation}

where $\mathcal{P}_{\delta}$ denotes the set of Pareto-efficient configurations satisfying the MAE tolerance constraint. The final model is then retrained on the full training set using the hyperparameters corresponding to $\alpha^{*}$ and evaluated once on the held-out test set, ensuring that model selection reflects an explicit and principled accuracy--fairness trade-off rather than an arbitrary choice of fairness weight.
\section{Experimental Setup}
\label{sec:experimental_setup}

All experiments are conducted on the official FairCVdb train--test split. All models are trained on multi-modal inputs consisting of standardised structured features and biography embeddings (original or scrubbed). Biography embeddings are pre-computed and cached before training.

For all experiments, we use a mid-fusion neural network architecture to combine structured CV features and biography embeddings. The biography embedding is first processed through two dense layers with ReLU activations. The resulting biography representation is concatenated with the standardized structured features and passed through two dense shared layers to obtain a fused representation. For single-task gender classification, this representation is connected to a binary classification output and trained with a BCE loss. For single-task score prediction, it is connected to a sigmoid regression output and trained to predict the gender-biased suitability score $T^G$ using MAE loss. For multi-task adversarial learning, the fused representation is shared by two task-specific heads: a regression head for predicting $T^G$ and a gender-classification head connected through a gradient reversal layer. The multi-task model is trained with the weighted combination of MAE and BCE losses in Eq.~\ref{eq:mt_loss}.

Hyperparameters are optimized using Optuna \cite{optuna} with the TPE sampler. For score-prediction experiments, model selection follows the Pareto-front-based procedure described in Section~\ref{pareto}. For each value of $\alpha$, hyperparameters are optimized independently over 50 trials. We tune learning rate, dropout, batch size, number of epochs, and hidden-layer sizes; for the multi-task model, we additionally tune the adversarial strength $\lambda$. We search over a grid of fairness weights $\alpha \in \mathcal{A}$, with values ranging from $0$ to $5$ across all experiments. The acceptable MAE degradation threshold is set to $\delta=0.3$ for original biographies and $\delta=0.1$ for scrubbed biographies. A larger tolerance is used for original biographies because they contain stronger gender cues and therefore require a wider accuracy--fairness trade-off region, while scrubbed biographies are already partially debiased and use a stricter tolerance.

\input{ST_cls}

\section{Results and Discussion}

Table~\ref{tab:ST_cls} shows that gender is highly predictable from original biographies across all embedding models, as expected. After gender scrubbing, accuracy drops substantially to 0.745--0.802. However, performance remains clearly above random guessing, indicating that scrubbed biographies still contain implicit gender cues that can be captured by the embeddings. Among the scrubbed results, \texttt{openai-text-embedding-3-large} gives the highest gender prediction accuracy, possibly due to its rich, high-dimensional semantic embeddings. In addition, male and female recall values are not perfectly balanced across embedding models. Some models recover male examples better, while others recover female examples better. This suggests that the remaining gender cues are not captured uniformly across embedding models. Since the imbalance is not consistently biased toward one gender group, it may reflect that the embedding models themselves encode gender-related associations differently, potentially producing biased representations after explicit gender indicators are removed. This is consistent with prior work showing that static word embeddings can exhibit different levels of social and gender bias depending on the model, training setting, bias dimension, and evaluation metric~\cite{biasedembedding,wefe}.

Comparing the results of single-task score prediction with multi-task adversarial learning on original biographies in Table \ref{tab:st_pareto_original_bios} and Table \ref{tab:mt_pareto_original_bios}, MAE decreases for most embedding models in the multi-task setting; however, since MAE is computed against the gender-biased scores $T^G$, a lower MAE does not necessarily indicate a higher unbiased prediction performance. We observe a consistent improvement in fairness-related metrics. For all embedding models, KL divergence decreases and p-percent increases, indicating that the predicted score distributions become more balanced across gender groups. The ranking-based metrics also improve: Average Recall@100 increases for all embeddings, and Female Recall@100 improves consistently. Although Male Recall@100 decreases slightly for some models, it remains relatively high across all embeddings. Most importantly, Recall Gap@100 as another fairness metric, decreases for every embedding model, showing that adversarial learning reduces the disparity between male and female recovery rates among the truly qualified candidates. Among the embedding models, \texttt{FacebookAI/roberta-base} provides the strongest overall trade-off in the multi-task setting with original biographies. It is among the top-performing models for four main metrics: KL divergence, p-percent, Average Recall@100, and Recall Gap@100 offering a particularly effective balance between predictive utility and fairness in this setting.

For scrubbed biographies, the results in Table~\ref{tab:st_pareto_scrubbed_bios} and Table~\ref{tab:mt_pareto_scrubbed_bios} show that the effect of multi-task adversarial learning is less uniform than in the original-biography setting. In terms of prediction performance, MAE decreases and Average Recall@100 increases for almost all embedding models. However, the fairness metrics do not improve consistently: for some embeddings the single-task model achieves lower KL divergence or Recall Gap@100, while for others the multi-task model performs better. Therefore, we cannot conclude that adversarial learning always improves fairness when the input already contains only implicit gender cues. Instead, its effect appears to depend strongly on the embedding model, which is consistent with the single-task gender classification results in Table \ref{tab:ST_cls} showing that different embedding models retain and encode residual gender-related information differently after scrubbing, often dissimilarly biased across gender groups. Among scrubbed biographies, \texttt{microsoft/mpnet-base} 
in the multi-task adversarial setting provides the strongest 
overall balance between predictive utility and fairness, 
achieving competitive MAE alongside top-three performance 
on KL divergence, P-percent, Average Recall@100, and 
Recall Gap@100.
\input{ST_pareto_orig}
\input{MT_pareto_orig}

Finally, comparing multi-task adversarial learning on original biographies in Table~\ref{tab:mt_pareto_original_bios} with single-task score prediction on scrubbed biographies in Table~\ref{tab:st_pareto_scrubbed_bios} shows that preprocessing-stage gender scrubbing leads to stronger fairness improvements; while Average Recall@100 of all embedding models remains in a comparable range for both, the scrubbed single-task setting achieves lower KL divergence and higher p-percent than the multi-task adversarial model trained on original biographies. Recall Gap@100 is also lower for almost all embeddings. This suggests that although adversarial learning reduces the effect of explicit gender information to some extent, it does not suppress gender-related cues as effectively as directly removing them from the text.

\input{ST_pareto_scrubbed}
\input{MT_pareto_scrubbed}

\section{Conclusion}

In this paper, we evaluated nine pre-trained text embedding models for gender bias in ML-based recruitment scoring, comparing single-task and multi-task adversarial learning with gradient reversal on both original and gender-scrubbed biographies. Our results show that all embedding models encode substantial gender information, and that gender scrubbing alone does not fully eliminate implicit gender cues. Multi-task adversarial learning consistently improves fairness on original biographies, but its benefit on scrubbed biographies varies across embedding models, reflecting differences in how residual gender information is encoded after explicit indicator removal. The choice of embedding model plays an important role in both predictive utility and fairness, and practitioners should carefully select embedding models for fairness-critical recruitment applications. Overall, preprocessing-stage gender scrubbing leads to stronger and more consistent fairness improvements, suggesting that adversarial training is best treated as a complementary strategy rather than a substitute for text-level debiasing.

This study comes with a few limitations that are left for future work. First, our study relies exclusively on FairCVdb, a synthetic dataset. Real CVs are noisier and contain a wider, less predictable, range of gender proxies. While we believe the overall conclusions will hold, validation on real recruitment data remains necessary before operational use.
Second, while we focus on binary gender bias, other types of bias, such as related to ethnicity, are also common in ML-based recruitment. The framework extends naturally to other protected attributes where intersectional groups could be addressed either through a multi-class adversary or through several parallel adversarial heads. Intersectional evaluation does, however, reduce per-group sample sizes, making top-100 ranking metrics noisier, and demographic-parity measures would need to be generalized beyond two groups.

\begin{acknowledgments}
We would like to thank the Flemish Government under the Onderzoeksprogramma Artificiele Intelligentie (AI) Vlaanderen programme for funding this research.
\end{acknowledgments}

\section*{Declaration on Generative AI}

The authors used GPT-5.5 to improve the language and readability of the manuscript. The authors reviewed and edited the content as needed and take full responsibility for the content of the publication.

\bibliography{bib}

\appendix

\end{document}

%% file: ST_cls.tex

\begin{table*}[t]
\centering
\caption{Gender prediction results across embedding models using original and scrubbed biographies.}
\label{tab:ST_cls}
\resizebox{\textwidth}{!}{
\begin{tabular}{l|ccc|ccc}
\hline
& \multicolumn{3}{c|}{\textbf{Original biographies}}
& \multicolumn{3}{c}{\textbf{Scrubbed biographies}} \\
\textbf{Embedding model}
& \textbf{Accuracy}
& \textbf{M Recall}
& \textbf{F Recall}
& \textbf{Accuracy}
& \textbf{M Recall}
& \textbf{F Recall} \\
\hline
fastText & 0.985 & 0.986 & 0.983 & 0.745 & 0.799 & 0.689 \\
GloVe & 0.976 & 0.974 & 0.979 & 0.757 & 0.727 & 0.788 \\
bert-base-uncased & 0.990 & 0.995 & 0.983 & 0.774 & 0.778 & 0.769\\
FacebookAI/roberta-base & 0.969 & 0.976 & 0.961 & 0.768 & 0.771 & 0.765 \\
microsoft/mpnet-base & 0.993 & 0.992 & 0.993 & 0.770 & 0.810 & 0.729 \\
all-mpnet-base-v2 & 0.977 & 0.977 & 0.977 & 0.754 & 0.738 & 0.770 \\
all-MiniLM-L6-v2 & 0.984 & 0.986 & 0.981 & 0.749 & 0.718 & 0.781 \\
openai-text-embedding-3-large & 0.995 & 0.996 & 0.994 & 0.802 & 0.830 & 0.774 \\
openai-text-embedding-3-small & 0.992 & 0.995 & 0.988 & 0.766 & 0.766 & 0.766 \\
\hline
\end{tabular}
}
\end{table*}

%% file: ST_pareto_orig.tex
\begin{table*}[t]
\centering
\caption{Single-task score prediction results on the test set using original biographies. \small\textbf{Bold} values indicate the top-three results per main metric.}
\label{tab:st_pareto_original_bios}
\resizebox{\textwidth}{!}{
\begin{tabular}{l|cccccccc}
\hline
\textbf{Embedding model}
& \textbf{$\alpha$}
& \textbf{MAE}
& \textbf{KL}
& \textbf{P\%}
& \multicolumn{4}{c}{\textbf{Recall@100}} \\
\cline{6-9}
&
&
&
&
& \textbf{Male}
& \textbf{Female}
& \textbf{Avg}
& \textbf{Gap} \\
\hline
fastText & 0.20 & 0.064 & 0.249 & 0.266 & 0.980 & 0.412 & \textbf{0.696} & 0.568 \\
GloVe & 0.10 & \textbf{0.029} & 0.248 & 0.235 & 0.980 & 0.353 & 0.667 & 0.627 \\
bert-base-uncased & 0.20 & \textbf{0.039} & \textbf{0.217} & 0.266 & 0.918 & 0.392 & 0.655 & \textbf{0.526} \\
FacebookAI/roberta-base & 0.30 & 0.058 & \textbf{0.178} & \textbf{0.299} & 0.898 & 0.353 & 0.625 & 0.545 \\
microsoft/mpnet-base & 0.20 & \textbf{0.044} & 0.273 & 0.250 & 0.918 & 0.373 & 0.646 & 0.545 \\
all-mpnet-base-v2 & 0.50 & 0.077 & 0.235 & \textbf{0.299} & 0.898 & 0.412 & 0.655 & \textbf{0.486} \\
all-MiniLM-L6-v2 & 0.20 & 0.064 & \textbf{0.163} & \textbf{0.316} & 0.918 & 0.392 & 0.655 & \textbf{0.526} \\
openai-text-embedding-3-large & 0.30 & 0.074 & 0.221 & \textbf{0.333} & 0.959 & 0.470 & \textbf{0.714} & \textbf{0.488} \\
openai-text-embedding-3-small & 0.50 & 0.062 & 0.261 & 0.282 & 0.939 & 0.412 & \textbf{0.675} & 0.527 \\
\hline
\end{tabular}
}
\end{table*}

%% file: MT_pareto_orig.tex
\begin{table*}[t]
\centering
\caption{Multi-task adversarial learning results on the test set with Pareto-front selection using original biographies. \small\textbf{Bold} values indicate the top-three results per main metric.}
\label{tab:mt_pareto_original_bios}
\resizebox{\textwidth}{!}{
\begin{tabular}{l|cccccccc}
\hline
\textbf{Embedding model}
& \textbf{$\alpha$}
& \textbf{MAE}
& \textbf{KL}
& \textbf{P\%}
& \multicolumn{4}{c}{\textbf{Recall@100}} \\
\cline{6-9}
&
&
&
&
& \textbf{Male}
& \textbf{Female}
& \textbf{Avg}
& \textbf{Gap} \\
\hline
fastText & 0.50 & 0.047 & \textbf{0.042} & 0.562 & 0.939 & 0.608 & 0.773 & 0.331 \\
GloVe & 0.30 & 0.042 & 0.082 & \textbf{0.695} & 0.857 & 0.667 & 0.762 & \textbf{0.190} \\
bert-base-uncased & 0.10 & \textbf{0.034} & 0.131 & 0.351 & 0.939 & 0.490 & 0.714 & 0.486 \\
FacebookAI/roberta-base & 0.20 & 0.047 & \textbf{0.034} & \textbf{0.667} & 0.898 & 0.686 & \textbf{0.792} & \textbf{0.212} \\
microsoft/mpnet-base & 0.10 & 0.036 & 0.100 & 0.587 & 0.898 & 0.647 & 0.773 & \textbf{0.251} \\
all-mpnet-base-v2 & 0.10 & 0.038 & 0.084 & 0.562 & 0.918 & 0.647 & \textbf{0.783} & 0.271 \\
all-MiniLM-L6-v2 & 0.20 & 0.044 & \textbf{0.055} & \textbf{0.613} & 0.857 & 0.549 & 0.703 & 0.308 \\
openai-text-embedding-3-large & 0.10 & \textbf{0.031} & 0.137 & 0.449 & 0.959 & 0.569 & 0.764 & 0.391 \\
openai-text-embedding-3-small & 0.10 & \textbf{0.030} & 0.160 & 0.493 & 0.959 & 0.608 & \textbf{0.784} & 0.351 \\
\hline
\end{tabular}
}
\end{table*}

%% file: ST_pareto_scrubbed.tex
\begin{table*}[t]
\centering
\caption{Single-task learning results on the test set with Pareto-front selection using scrubbed biographies. \small\textbf{Bold} values indicate the top-three results per main metric.}
\label{tab:st_pareto_scrubbed_bios}
\resizebox{\textwidth}{!}{
\begin{tabular}{l|cccccccc}
\hline
\textbf{Embedding model}
& \textbf{$\alpha$}
& \textbf{MAE}
& \textbf{KL}
& \textbf{P\%}
& \multicolumn{4}{c}{\textbf{Recall@100}} \\
\cline{6-9}
&
&
&
&
& \textbf{Male}
& \textbf{Female}
& \textbf{Avg}
& \textbf{Gap} \\
\hline
fastText & 0.10 & \textbf{0.051} & 0.022 & 0.724 & 0.837 & 0.647 & \textbf{0.742} & 0.190 \\
GloVe & 2.00 & 0.053 & \textbf{0.014} & \textbf{0.754} & 0.816 & 0.608 & 0.712 & 0.208 \\
bert-base-uncased & 0.50 & \textbf{0.050} & 0.023 & 0.724 & 0.816 & 0.667 & \textbf{0.741} & 0.150 \\
FacebookAI/roberta-base & 1.00 & 0.058 & 0.023 & 0.724 & 0.796 & 0.667 & 0.731 & \textbf{0.129} \\
microsoft/mpnet-base & 1.00 & 0.054 & \textbf{0.012} & 0.695 & 0.816 & 0.667 & \textbf{0.741} & 0.150 \\
all-mpnet-base-v2 & 1.00 & 0.052 & 0.028 & 0.724 & 0.796 & 0.667 & 0.731 & \textbf{0.129} \\
all-MiniLM-L6-v2 & 0.50 & \textbf{0.051} & \textbf{0.021} & 0.667 & 0.837 & 0.627 & 0.732 & 0.209 \\
openai-text-embedding-3-large & 1.00 & 0.087 & 0.031 & \textbf{0.852} & 0.857 & 0.706 & \textbf{0.782} & 0.151 \\
openai-text-embedding-3-small & 2.00 & 0.082 & 0.025 & \textbf{0.923} & 0.776 & 0.686 & 0.731 & \textbf{0.089} \\
\hline
\end{tabular}
}
\end{table*}

%% file: MT_pareto_scrubbed.tex
\begin{table*}[t]
\centering
\caption{Multi-task adversarial learning results on the test set with Pareto-front selection using scrubbed biographies. \small\textbf{Bold} values indicate the top-three results per main metric.}
\label{tab:mt_pareto_scrubbed_bios}
\resizebox{\textwidth}{!}{
\begin{tabular}{l|cccccccc}
\hline
\textbf{Embedding model}
& \textbf{$\alpha$}
& \textbf{MAE}
& \textbf{KL}
& \textbf{P\%}
& \multicolumn{4}{c}{\textbf{Recall@100}} \\
\cline{6-9}
&
&
&
&
& \textbf{Male}
& \textbf{Female}
& \textbf{Avg}
& \textbf{Gap} \\
\hline
fastText & 1.00 & 0.051 & \textbf{0.015} & 0.724 & 0.837 & 0.667 & 0.752 & 0.170 \\
GloVe & 1.00 & 0.052 & \textbf{0.017} & \textbf{0.818} & 0.878 & 0.706 & 0.792 & 0.172 \\
bert-base-uncased & 1.00 & \textbf{0.048} & 0.021 & 0.754 & 0.918 & 0.686 & \textbf{0.802} & 0.232 \\
FacebookAI/roberta-base & 0.50 & 0.049 & 0.021 & 0.639 & 0.878 & 0.627 & 0.753 & 0.250 \\
microsoft/mpnet-base & 0.50 & 0.050 & \textbf{0.015} & \textbf{0.786} & 0.878 & 0.745 & \textbf{0.811} & \textbf{0.132} \\
all-mpnet-base-v2 & 0.50 & 0.050 & 0.021 & 0.754 & 0.918 & 0.686 & \textbf{0.802} & 0.232 \\
all-MiniLM-L6-v2 & 0.50 & 0.050 & 0.028 & \textbf{0.887} & 0.837 & 0.706 & 0.771 & \textbf{0.131} \\
openai-text-embedding-3-large & 0.10 & \textbf{0.043} & 0.048 & \textbf{0.786} & 0.837 & 0.725 & 0.781 & \textbf{0.111} \\
openai-text-embedding-3-small & 0.50 & \textbf{0.047} & 0.028 & 0.667 & 0.918 & 0.667 & \textbf{0.793} & 0.252 \\
\hline
\end{tabular}
}
\end{table*}